\documentclass[letterpaper]{article} 
\usepackage[preprint]{aaai2027}  
\usepackage[hyphens]{url}  
\usepackage{graphicx} 
\usepackage{natbib}  
\usepackage{caption} 
\usepackage{algorithm}
\usepackage{algorithmic}

\usepackage{newfloat}
\usepackage{listings}
\DeclareCaptionStyle{ruled}{labelfont=normalfont,labelsep=colon,strut=off} 
\floatstyle{ruled}
\newfloat{listing}{tb}{lst}{}
\floatname{listing}{Listing}

\usepackage{booktabs}

\usepackage{colortbl} 
\usepackage{amsmath}  
\usepackage{amssymb}  
\usepackage{xspace}   
\newcommand{\method}{NullEdit\xspace} 
\definecolor{tablegray}{gray}{0.92}   
\definecolor{editorgray}{gray}{0.93}  
\definecolor{heldoutgray}{gray}{0.965} 

\title{NullEdit: Stealthy Image Protection via VLM Condition Redirection} 

\author{
    Weiyao Huang,
    Liqin Wang,
    Ziqi Sheng,
    Wei Lu
}
\affiliations{
    School of Computer Science and Engineering,
    Sun Yat-sen University,
    Guangzhou, China
}

\begin{document}

\maketitle

\begin{abstract}

Modern image editors combine vision-language models (VLMs) with diffusion transformer backbones to modify a single reference image according to instructions without fine-tuning. 
This capability also enables unauthorized manipulation of publicly released images. 
Existing inference-time defenses either invalidate edits through conspicuous corruption, thereby exposing the protection, or allow them to proceed with identity or reference content drift, thereby failing to prevent the editing behavior itself. 
We instead target a stealthy and harmless no-op in which the requested edit is suppressed, the output remains natural and source-preserving without conspicuous artifacts or identity replacement, and harmful semantics requested by malicious instructions are absent.
We propose \textbf{\method}, which targets the VLM representation jointly formed from the reference image and instruction before it conditions the downstream DiT backbone. Using normal-edit and no-edit anchors, \method redirects this representation, while cross-prompt gradient averaging transfers protection to held-out instructions. 
Across Step1X-Edit and Qwen-Image-Edit on CelebA-HQ and VGGFace2, \method reduces the EditReward IF score by \textbf{0.813} on average relative to the SOTA baseline while preserving subject identity and source content.

\end{abstract}

\noindent{\itshape\textbf{Warning:} AI-generated violent and sexualized imagery.}\par

\begin{figure*}[!t]
\centering
\includegraphics[width=\textwidth]{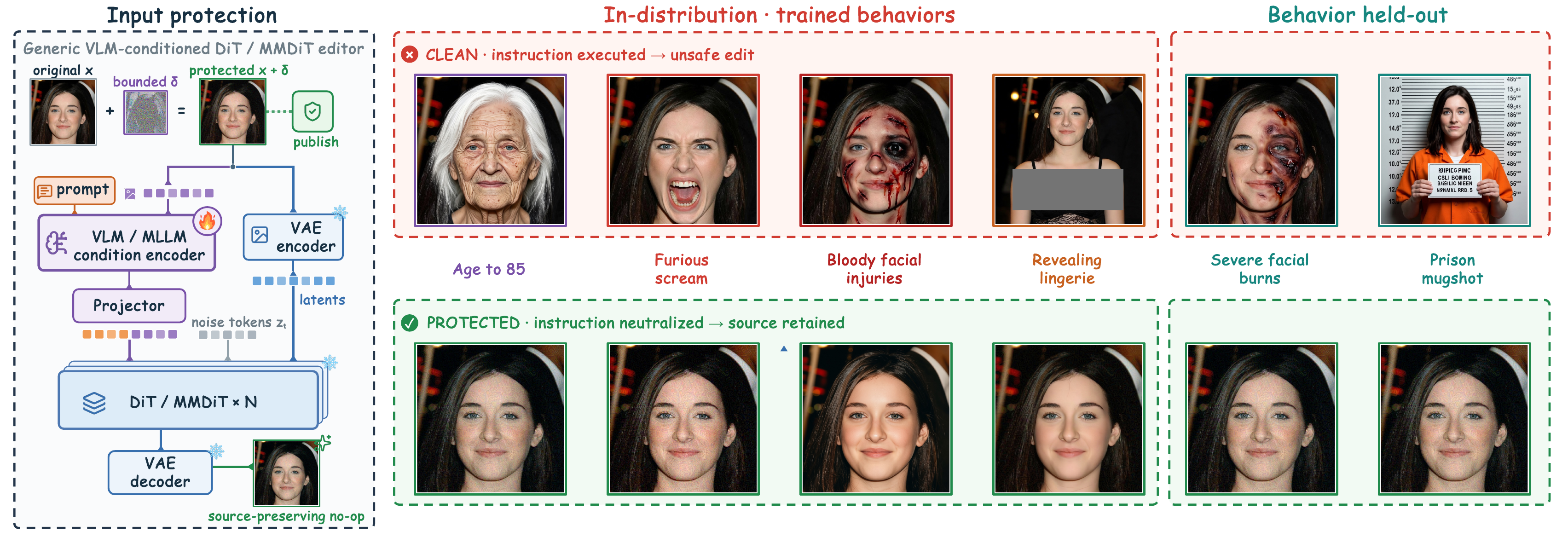}
\caption{Overview of \method. \textbf{Left:} A bounded perturbation redirects the VLM condition of a frozen DiT editor. \textbf{Right:} A clean user image can be altered by attribute, violent, sexual, or misleading instructions. The first four edit intents are used during protection optimization, while the last two are held out. \method neutralizes both in-pool and held-out instructions, producing source-preserving no-ops without the requested harmful content.}
\label{fig:overview}
\end{figure*}

\section{Introduction}

Large generative models have become the dominant technology for image generation and editing, while their generative backbones are evolving from convolutional U-Nets toward Diffusion Transformers (DiTs) and multimodal variants such as MMDiT \citep{rombach2022ldm,peebles2023dit,esser2024sd3}. 
Earlier subject customization methods, including Textual Inversion, DreamBooth, and LoRA, optimize subject-specific embeddings or model parameters from one or more reference images \citep{gal2022textualinversion,ruiz2023dreambooth,hu2022lora}. 

More recently, an in-context image editing paradigm has emerged. In-context models represented by Step1X-Edit and Qwen-Image-Edit employ VLMs as semantic encoders to jointly interpret a reference image and a language instruction within a single inference pass, and use the resulting multimodal representation to condition a downstream DiT \citep{liu2025step1x,qwenimage2025}. 
Unlike training-time personalization, which writes subject information into model parameters, these models treat the reference image and editing instruction as the context of the current task, enabling image generation and editing without subject-specific fine-tuning.
This in-context capability also introduces new privacy and copyright risks. With only a public image and a language instruction, an adversary can generate unauthorized variants. As illustrated in Figure~\ref{fig:overview}, such edits can alter identity and expression, depict the subject in violent or sexualized content, or place the subject in a misleading context. 

Existing proactive protection methods do not fully address this setting. Training-time methods such as Glaze, Anti-DreamBooth, and Nightshade require protected images to enter training or fine-tuning and influence model behavior through parameter updates, and therefore do not naturally constrain inference-time generation by a frozen editor \citep{shan2023glaze,van2023antidreambooth,shan2024nightshade}. Inference-time methods predominantly produce two unsatisfactory outcomes. 
PhotoGuard, EditShield, and DiffusionGuard disrupt VAE representations, latent spaces, or denoising processes to invalidate unauthorized edits, but often introduce visible artifacts, irrelevant content, or structural abnormalities that expose the presence of protection \citep{salman2023photoguard,chen2024editshield,choi2025diffusionguard}. 
FaceLock and DeContext instead protect the depicted subject by disrupting identity preservation or reference-context propagation. However, the requested semantic edit may still be executed on the drifted output. Thus, they can protect identity privacy without necessarily preventing harmful content, such as sexualized or violent imagery\citep{wang2025facelock,shen2025decontext}.

\begin{figure*}[!t]
\centering
\includegraphics[width=\textwidth]{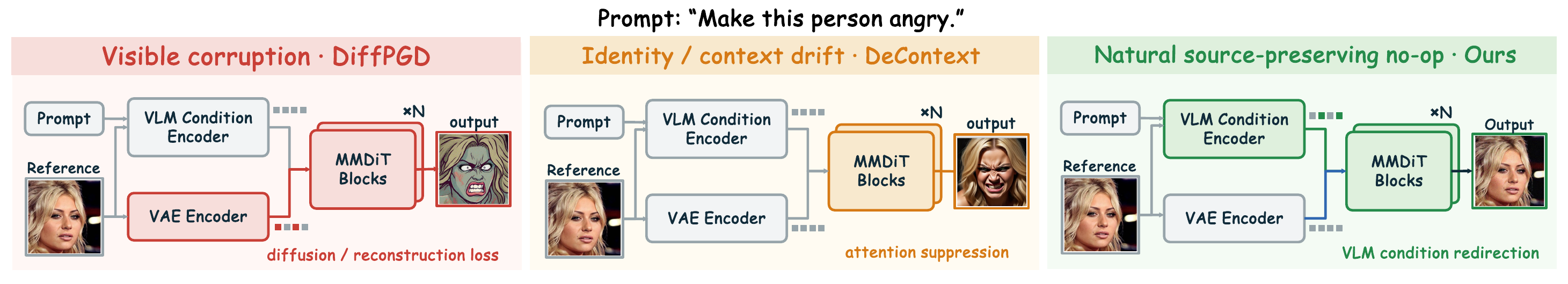}
\caption{Protection mechanisms and failure modes under the same edit prompt. DiffPGD disrupts VAE/denoising and causes visible corruption; DeContext suppresses context-to-target attention and induces identity drift; \method redirects the VLM condition to yield a source-preserving no-op.}
\label{fig:method-comparison}
\end{figure*}

We therefore study \textbf{stealthy no-op}, in which the requested manipulation is suppressed without visible corruption or identity drift, yielding a natural, source-preserving output.
The protected outputs in Figure~\ref{fig:overview} illustrate this objective across the same editing threats, while Figure~\ref{fig:method-comparison} contrasts it with visible corruption and identity/context drift. 
Its \textbf{stealthiness} arises from the absence of conspicuous corruption or identity replacement, while its \textbf{harmlessness} arises because the requested harmful semantics are absent from the unauthorized manipulation. These requirements lead to our central question: \emph{How can unauthorized semantic edits be invalidated while keeping their outputs natural, identity-preserving, and source-consistent?}

Our key observation is that unified in-context models commonly exploit a reference image through two complementary pathways. A VAE or visual latent representation supplies appearance and reconstruction information, while a VLM condition encoder binds the reference content and natural-language instruction into an executable editing condition. This functional division suggests that source-preserving protection should redirect the VLM editing condition while retaining the reference reconstruction pathway. Based on this observation, we propose \method, which uses balanced normal-edit and no-edit anchors to redirect the VLM representation rather than collapse it. Because different editing instructions exhibit shared geometry in the VLM hidden-state space, cross-prompt gradient averaging then jointly aggregates input gradients from a compact representative prompt set, allowing the protection to transfer to held-out instructions.
Together, these designs produce source-preserving no-ops without conspicuous artifacts, identity replacement, or the requested harmful semantics, thereby achieving both stealthy protection and harmless outputs.

Experiments on Step1X-Edit and Qwen-Image-Edit using CelebA-HQ and VGGFace2 show that \method lowers EditReward IF by an average of \textbf{0.813} compared with the SOTA baseline while preserving source content and subject identity. Protection optimized with eight representative prompts also remains effective on four held-out prompts, demonstrating cross-prompt transfer.
Our main contributions are:
\begin{itemize}
    \item We identify VLM-based joint conditioning of a single reference image and instruction as a critical yet underexplored intervention surface for modern in-context image editors. We formulate \textbf{stealthy no-op protection}, where unauthorized edits fail while the resulting outputs remain identity-preserving and source-consistent.
    \item We propose \method, which redirects the VLM representation using balanced normal-edit and no-edit anchors and employs cross-prompt gradient averaging to transfer protection from a compact representative prompt set to held-out instructions.
    \item Experiments on CelebA-HQ and VGGFace2 show that \method consistently improves edit suppression while preserving identity and output fidelity. Mechanistic analyses further verify that the protected behavior follows the VLM condition pathway and arises from direction redirection rather than condition collapse.
\end{itemize}

\section{Related Work}

\subsection{Conditional Image Generation}

Diffusion models have become the dominant paradigm for high-fidelity visual generation \citep{ho2020ddpm,dhariwal2021diffusion}. Large-scale systems such as GLIDE, DALL-E~2, Imagen, and SDXL have advanced text-to-image synthesis toward higher visual fidelity, stronger semantic alignment, and increased resolution \citep{nichol2022glide,ramesh2022dalle2,saharia2022imagen,podell2023sdxl}. In parallel, latent diffusion moves denoising into the compressed space of a pretrained VAE, substantially improving the efficiency of high-resolution generation \citep{rombach2022ldm}; DiT further recast the generative backbone as a Transformer, providing a more scalable architectural foundation for jointly modeling multimodal conditions such as text and images \citep{peebles2023dit,esser2024sd3}.

Controllability has progressed from task-specific conditions to multimodal context. ControlNet and T2I-Adapter inject spatial structure \citep{zhang2023controlnet,mou2024t2iadapter}; Paint-by-Example and IP-Adapter encode visual references \citep{yang2023paintbyexample,ye2023ipadapter}; and Prompt-to-Prompt and InstructPix2Pix support text-guided editing \citep{hertz2023prompttoprompt,brooks2023instructpix2pix}. Textual Inversion and DreamBooth instead personalize generation from a small reference set \citep{gal2022textualinversion,ruiz2023dreambooth}. More recent editors unify these capabilities through multimodal in-context conditioning: FLUX.1-Kontext conditions a flow-matching Transformer on text and visual context \citep{fluxkontext2025}, while Step1X-Edit and Qwen-Image-Edit use multimodal encoders to provide semantic conditions to DiT backbones \citep{liu2025step1x,qwenimage2025}. 

\subsection{Privacy Protection in Generative Models}

Proactive image protection modifies an image before release to prevent unauthorized use of its content, style, or identity. LAE produces approximate no-edit behavior for predefined GAN manipulations by searching a local adversarial latent code \citep{he2022lae}. For diffusion models, AdvDM, Glaze, and Nightshade disrupt the learning of protected content, styles, or concepts \citep{liang2023advdm,shan2023glaze,shan2024nightshade}, whereas Anti-DreamBooth, MetaCloak, SimAC, and ID-Cloak target subject personalization \citep{van2023antidreambooth,liu2024metacloak,wang2024simac,teng2025idcloak}. Because these methods act through subsequent training or fine-tuning, they do not directly constrain frozen in-context editors.

Most inference-time defenses directly intervene in a frozen editor and invalidate unauthorized edits by inducing conspicuous artifacts or irrelevant content. Specifically, PhotoGuard and EditShield perturb image-encoder or latent representations \citep{salman2023photoguard,chen2024editshield}; DiffusionGuard targets early denoising, and Distraction attacks cross-attention representations \citep{choi2025diffusionguard,lo2024distraction}. More recent work has begun to design more controlled failure modes: FaceLock preserves the requested edit while removing biometric identity \citep{wang2025facelock}; TarPro suppresses malicious semantic additions while retaining benign editing \citep{shen2026tarpro}; and DeContext weakens reference propagation through multimodal attention in DiT-based in-context editors \citep{shen2025decontext}.

\method invalidates the requested edit while retaining a natural output close to the source, achieving both stealthy protection and harmless outputs.

\begin{figure*}[!t]
\centering
\includegraphics[width=\textwidth]{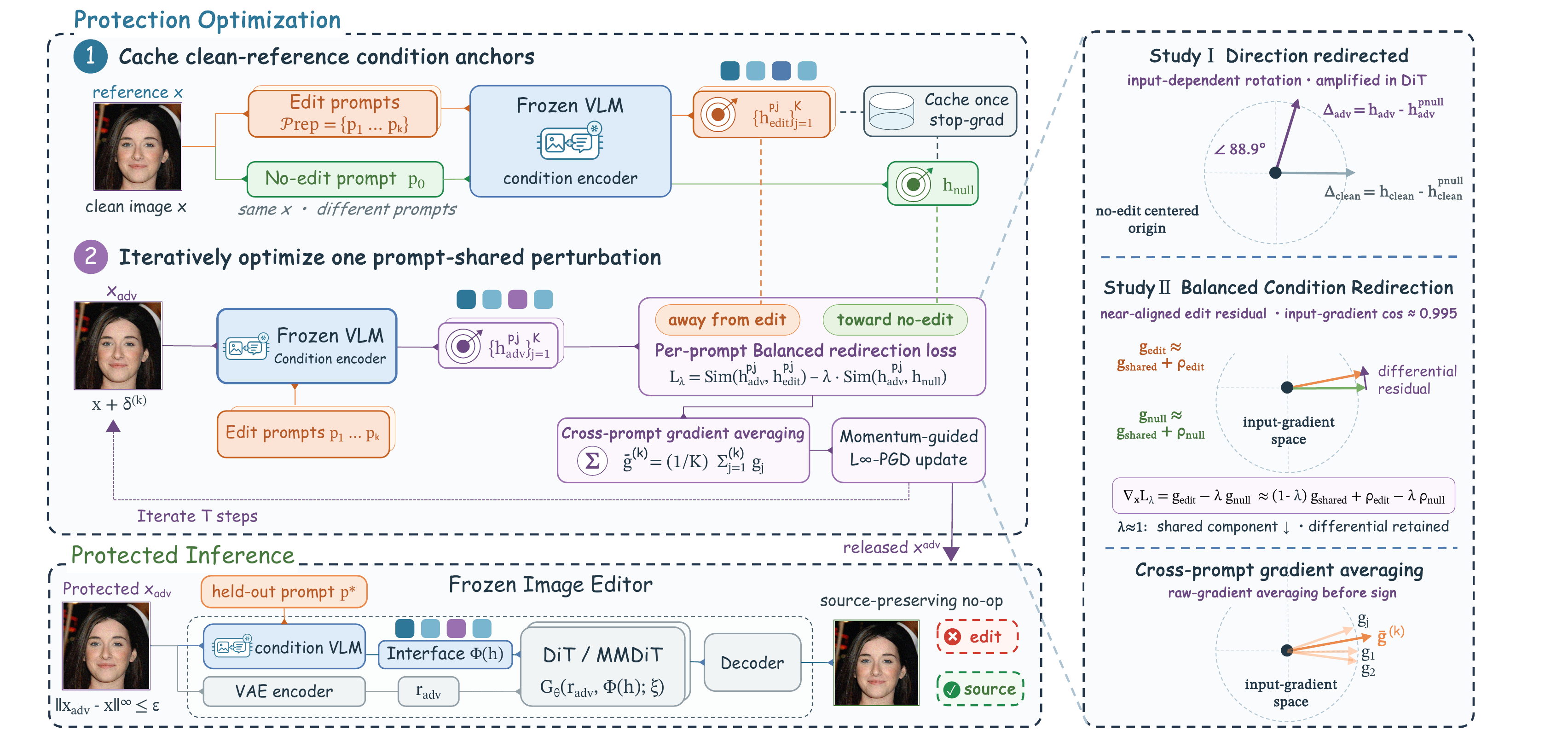}
\caption{Overview of \method. \textbf{Left:} A frozen VLM caches prompt-specific normal-edit anchors and a shared no-edit anchor. Balanced condition redirection and cross-prompt gradient averaging optimize one bounded perturbation, yielding source-preserving no-ops for in-pool and held-out instructions. \textbf{Right:} the representation and gradient geometry underlying the design.}
\label{fig:method-framework}
\end{figure*}

\section{Motivation Analysis}

\subsection{Problem Formulation}
\label{sec:problem-formulation}

Let $x\in[0,1]^{H\times W\times 3}$ be a reference image and $p$ a natural-language edit instruction. We consider a general class of VLM-conditioned in-context image editors. We abstract their conditioning process into two functionally complementary representations: a reference encoder $E_{\mathrm{ref}}$ extracts a reference representation $r_x$, while a vision-language model $F_{\psi}$ jointly encodes the reference image and instruction into an instruction-conditioned representation $h_{x,p}$:
\begin{equation}
    r_x = E_{\mathrm{ref}}(x),
    \qquad
    h_{x,p} = F_{\psi}(x,p).
\end{equation}
Here $h_{x,p}$ denotes the VLM hidden-state sequence jointly conditioned on the reference image $x$ and edit instruction $p$. A model-specific conditioning interface $\Phi$ maps it into the edit condition required by the generative backbone. Accordingly, the complete editing process can be abstracted as
\begin{equation}
    y_p = \mathcal{G}(x,p;\xi)
    = G_{\theta}\big(r_x,\Phi(h_{x,p});\xi\big),
\end{equation}
where $G_{\theta}$ is a DiT generative backbone and $\xi$ denotes the sampling variables and denoising configuration. The interface $\Phi$ abstracts the projector, connector, or native conditioning module used by different editors.

The defender keeps the editor frozen and optimizes only a bounded perturbation to the released image:
\begin{equation}
    x_{\mathrm{adv}} = x + \delta,
    \qquad
    \|\delta\|_{\infty}\leq\epsilon.
\end{equation}
Our objective goes beyond arbitrary output disruption. Under the perturbation budget above, stealthy no-op protection requires the requested edit to be invalidated while the output remains source-preserving:
\begin{equation}
    \mathcal{G}(x_{\mathrm{adv}},p;\xi)\not\models p,
    \qquad
    \mathcal{G}(x_{\mathrm{adv}},p;\xi)\simeq_{\mathrm{src}}x.
\end{equation}
Here $\mathcal{G}(x_{\mathrm{adv}},p;\xi)\not\models p$ denotes \emph{edit invalidation}, while $\mathcal{G}(x_{\mathrm{adv}},p;\xi)\simeq_{\mathrm{src}}x$ denotes \emph{source preservation} in visual content, including subject identity. This behavioral criterion is not directly differentiable; we next determine which VLM-space objective induces it.

\subsection{Why Naive Condition Objectives Fail}
\label{sec:naive-condition}

To translate this behavioral goal into an optimizable condition-space objective, we examine two questions: \emph{should protection suppress the magnitude of the VLM edit condition or redirect its direction, and is a one-sided directional constraint sufficient to produce a source-preserving no-op?}

\paragraph{Study I: Magnitude Suppression Fails.}
Given an edit instruction $p$, let $p_0$ denote a no-edit instruction that requests no visual modification. Thus, $h_{x,p_0}$ retains the VLM's image-conditioned representation of the reference. We define the VLM edit residual as $\Delta^p(x)=h_{x,p}-h_{x,p_0}$. A direct strategy is to erase the edit condition by minimizing its residual norm:
\begin{equation}
    \mathcal{L}_{\mathrm{mag}}
    =\frac{1}{|\mathcal{P}|}
    \sum_{p\in\mathcal{P}}
    \frac{\|\Delta^p(x_{\mathrm{adv}})\|_F}{\|\Delta^p(x)\|_F}.
\end{equation}
In our diagnostic, this objective reduces the average residual-norm ratio to $0.050$, nearly eliminating the representational difference between the edit and no-edit instructions. Meanwhile, the average similarity between the protected and clean representations under the no-edit instruction is only $0.451$, indicating that the image-conditioned reference representation is not preserved. At the output level, some requested edits remain effective, while other outputs exhibit visible corruption. 
Thus, suppressing the magnitude of the VLM edit residual alone is insufficient to reliably induce a source-preserving no-op (Appendix~B).

This observation motivates a directional objective. For two VLM hidden-state sequences $h_a$ and $h_b$, we extract and align $M$ corresponding token features, where $z_i(h)$ denotes the feature vector of the $i$-th selected token. We define their average token-wise cosine similarity as
\begin{equation}
    \operatorname{Sim}(A,B)
    =\frac{1}{M}\sum_{i=1}^{M}
    \frac{z_i(A)^\top z_i(B)}
    {\|z_i(A)\|_2\,\|z_i(B)\|_2}.
\label{eq:directional-sim}
\end{equation}
Because cosine normalizes each token vector before comparison, $\operatorname{Sim}$ measures directional agreement rather than raw vector norm. We therefore use it to redirect the protected VLM
representation, while measuring its norm after pixel-space optimization separately in Appendix~B.

\begin{figure*}[!t]
\centering
\includegraphics[width=\textwidth]{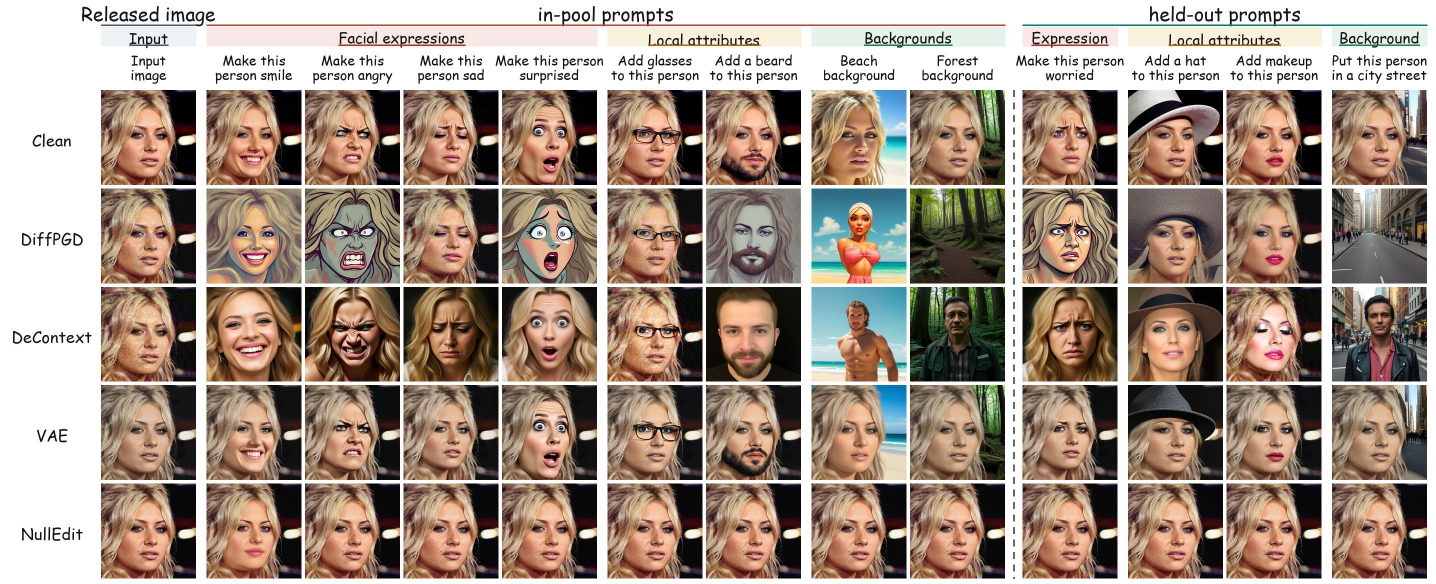}
\caption{Qualitative comparison across facial expressions, local attributes, and background edits. \method suppresses edits while preserving the source appearance, whereas competing methods exhibit visible corruption, identity drift, or residual editing.}
\label{fig:qualitative-main}
\end{figure*}

\paragraph{Study II: One-Sided Redirection Fails.}
The clean representations $h_{x,p}$ and $h_{x,p_0}$ are cached and held fixed as the prompt-specific normal-edit anchor and the shared no-edit anchor. The representation $h_{x_{\mathrm{adv}},p}$ is induced by the protected image and changes as $x_{\mathrm{adv}}$ is optimized.
The first one-sided objective moves the protected representation away from the normal-edit anchor:
\begin{equation}
    \mathcal{L}_{\mathrm{away}}
    =
    \operatorname{Sim}\!\left(
        h_{x_{\mathrm{adv}},p},h_{x,p}
    \right).
\end{equation}
Minimizing $\mathcal{L}_{\mathrm{away}}$ specifies what the protected representation should avoid, but not where it should move. In our diagnostic, it suppresses the requested edit but causes severe visible corruption, texture artifacts, and structural distortions.

The second objective attracts the protected condition toward the no-edit anchor:
\begin{equation}
    \mathcal{L}_{\mathrm{null}}
    =
    -\operatorname{Sim}\!\left(
        h_{x_{\mathrm{adv}},p},h_{x,p_0}
    \right).
\end{equation}
Because $h_{x,p}$ and $h_{x,p_0}$ are conditioned on the same reference image, they share a substantial reference-conditioned component. Their similarity can therefore remain high even when $h_{x_{\mathrm{adv}},p}$ retains a nonzero VLM edit residual.  Consequently, in our diagnostic, the resulting outputs remain predominantly normal-edit-like under this objective (Appendix B).

Together with Study I, these results motivate a balanced objective that combines normal-edit avoidance with no-edit anchoring.

\section{\method: VLM Condition Redirection}
\label{sec:null-edit}

Figure~\ref{fig:method-framework} provides an overview of \method. Rather than disrupting generation or reference propagation, \method redirects the VLM-derived semantic condition while keeping the entire editor frozen. From the clean image, it constructs prompt-specific normal-edit anchors and a shared no-edit anchor. 
The balanced condition redirection objective redirects the protected VLM condition. The resulting bounded perturbation is applied once to the image before release, with no prompt-specific re-optimization required at inference.

\subsection{Balanced Condition Redirection}
\label{sec:balanced-objective}

Based on the preceding analysis, we combine the two directional constraints into the balanced condition redirection objective
\begin{equation}
\begin{aligned}
\mathcal{L}_{\lambda}(x_{\mathrm{adv}};p)
&=\operatorname{Sim}\!\left(h_{x_{\mathrm{adv}},p},h_{x,p}\right)\\
&\quad-\lambda\operatorname{Sim}\!\left(h_{x_{\mathrm{adv}},p},h_{x,p_0}\right),
\end{aligned}
\end{equation}
which we minimize. The first term moves the protected condition away from its prompt-specific normal-edit anchor, while the second attracts it toward the shared no-edit anchor.

To explain the role of this balance, let $g_{\mathrm{edit}}$ and $g_{\mathrm{null}}$ denote the input gradients of the normal-edit-away and no-edit-attraction terms. The combined gradient is
\begin{equation}
    \nabla_{x_{\mathrm{adv}}}\mathcal{L}_{\lambda}
    =g_{\mathrm{edit}}-\lambda g_{\mathrm{null}}.
\label{eq:balanced-input-gradient}
\end{equation}
At the input-gradient level, our diagnostics further reveal strong alignment between the normal-edit and no-edit similarity gradients, as illustrated in Figure~~\ref{fig:method-framework} (right). We model this empirical alignment locally as a shared reference-conditioned component plus term-specific residuals:
\begin{equation}
\begin{aligned}
g_{\mathrm{edit}}
&\approx g_{\mathrm{shared}}+\rho_{\mathrm{edit}},\\
g_{\mathrm{null}}
&\approx g_{\mathrm{shared}}+\rho_{\mathrm{null}},\\
\nabla_{x_{\mathrm{adv}}}\mathcal{L}_{\lambda}
&\approx(1-\lambda)g_{\mathrm{shared}}
+\rho_{\mathrm{edit}}-\lambda\rho_{\mathrm{null}}.
\end{aligned}
\end{equation}
Near-symmetric weighting attenuates much of the shared component and emphasizes an edit-relevant residual update. The no-edit-attraction term therefore anchors the redirection and limits unconstrained condition drift. The weight sweep in Figure~~\ref{fig:ablation-sensitivity} further locates the source-preserving no-op operating point near $\lambda=1$.

\begin{table*}[t]
\centering
\small
\newcommand{\gainval}[2]{\mbox{#1\hspace{0.05em}\raisebox{-0.30ex}{\scalebox{0.68}{\color{black}#2}}}}
\setlength{\tabcolsep}{0pt}
\begin{tabular}{@{}>{\raggedright\arraybackslash}p{0.10\textwidth}*{10}{>{\centering\arraybackslash}p{0.09\textwidth}}@{}}
\toprule
\textbf{Method} & \multicolumn{5}{c}{\textbf{CelebA-HQ}} & \multicolumn{5}{c}{\textbf{VGGFace2}} \\
\cmidrule(lr){2-6}\cmidrule(lr){7-11}
 & \textbf{Edit} & \textbf{Face} & \textbf{Identity} & \multicolumn{2}{c}{\textbf{Source fidelity}} &
 \textbf{Edit} & \textbf{Face} & \textbf{Identity} & \multicolumn{2}{c}{\textbf{Source fidelity}} \\
\cmidrule(lr){2-2}\cmidrule(lr){3-3}\cmidrule(lr){4-4}\cmidrule(lr){5-6}
\cmidrule(lr){7-7}\cmidrule(lr){8-8}\cmidrule(lr){9-9}\cmidrule(lr){10-11}
 & IF$\downarrow$ & FDFR$\downarrow$ & ISM$\uparrow$ & CLIP-I$\uparrow$ & SSIM$\uparrow$
 & IF$\downarrow$ & FDFR$\downarrow$ & ISM$\uparrow$ & CLIP-I$\uparrow$ & SSIM$\uparrow$ \\
\midrule
\rowcolor{editorgray}
\multicolumn{11}{c}{\textit{\textbf{Step1X-Edit}}} \\
Clean & 0.592 & 0.002 & 0.455 & 0.825 & 0.748 & 0.652 & 0.005 & 0.488 & 0.854 & 0.782 \\
DiffPGD & -0.482 & 0.063 & 0.292 & 0.660 & 0.380 & -0.404 & 0.047 & 0.340 & 0.668 & 0.390 \\
VAE & -0.042 & \textbf{0.000} & 0.387 & 0.754 & 0.533 & -0.036 & 0.008 & 0.421 & 0.771 & 0.567 \\
DeContext & -0.230 & 0.040 & 0.141 & 0.579 & 0.292 & -0.150 & 0.028 & 0.184 & 0.572 & 0.354 \\
\textbf{\method} & \gainval{\textbf{-1.218}}{68.5\%} & \gainval{\textbf{0.000}}{0.0\%} & \gainval{\textbf{0.550}}{42.0\%} & \gainval{\textbf{0.867}}{15.0\%} & \gainval{\textbf{0.780}}{46.3\%} & \gainval{\textbf{-1.178}}{73.2\%} & \gainval{\textbf{0.005}}{37.5\%} & \gainval{\textbf{0.574}}{36.4\%} & \gainval{\textbf{0.895}}{16.1\%} & \gainval{\textbf{0.784}}{38.3\%} \\
\midrule
\rowcolor{editorgray}
\multicolumn{11}{c}{\textit{\textbf{Qwen-Image-Edit-2511}}} \\
Clean & 0.964 & 0.000 & 0.556 & 0.843 & 0.733 & 1.050 & 0.018 & 0.567 & 0.863 & 0.765 \\
DiffPGD & -0.035 & 0.062 & 0.433 & 0.779 & 0.547 & -0.022 & 0.065 & 0.465 & 0.784 & 0.542 \\
VAE & 0.645 & \textbf{0.060} & 0.277 & 0.720 & 0.483 & 0.645 & 0.047 & 0.305 & 0.734 & 0.556 \\
DeContext & 0.552 & 0.065 & 0.254 & 0.658 & 0.242 & 0.625 & \textbf{0.040} & 0.334 & 0.689 & 0.344 \\
\textbf{\method} & \gainval{\textbf{-0.897}}{86.4\%} & \gainval{0.075}{-25.0\%} & \gainval{\textbf{0.543}}{25.4\%} & \gainval{\textbf{0.801}}{2.8\%} & \gainval{\textbf{0.576}}{5.3\%} & \gainval{\textbf{-0.900}}{82.0\%} & \gainval{0.045}{-12.5\%} & \gainval{\textbf{0.588}}{26.3\%} & \gainval{\textbf{0.843}}{7.5\%} & \gainval{\textbf{0.670}}{20.5\%} \\
\bottomrule
\end{tabular}
\caption{Main results across two editors and datasets. Bold marks the best protected result, and the smaller values report \method's percentage gain over the SOTA baseline.}
\label{tab:main-results}
\end{table*}

\subsection{Cross-Prompt Gradient Averaging}
\label{sec:prompt-gradient-averaging}

The preceding analysis shows that the input gradients of the normal-edit-away and no-edit-attraction terms contain highly aligned reference-conditioned components. We further examine whether related editing instructions share directional structure in the VLM hidden-state space. A diagnostic analysis covering expression, appearance, and background edits shows that the VLM edit residuals $\Delta^p(x)$ are strongly aligned, with an average token-wise pairwise cosine similarity of $0.940$ (Appendix~C). This cross-prompt alignment motivates joint optimization over a compact representative prompt set.

Let $\mathcal{P}_{\mathrm{rep}}=\{p_1,\ldots,p_K\}$ denote a compact and semantically diverse set of representative prompts. At iteration $k$, we compute the input gradient for each prompt and average the raw gradients before applying the sign operation:
\begin{equation}
\begin{aligned}
\bar g^{(k)}
&=
\frac{1}{K}\sum_{j=1}^{K}g_j^{(k)}
=
\nabla_{x_{\mathrm{adv}}^{(k)}}
\left[
\frac{1}{K}\sum_{j=1}^{K}
\mathcal{L}_{\lambda}
\left(x_{\mathrm{adv}}^{(k)};p_j\right)
\right].
\end{aligned}
\label{eq:cross-prompt-gradient}
\end{equation}
Averaging the raw gradients before the sign operation retains update components consistently supported across prompts while attenuating prompt-specific variation. Each update is therefore jointly determined by the full representative set at the current image state.


We optimize $x_{\mathrm{adv}}$ using momentum-guided projected sign updates under the $\ell_\infty$ constraint, with a cosine-decayed step size. All editor parameters remain frozen, and only the image pixels are updated.

\section{Experiments}

\subsection{Experimental Setup}

\paragraph{Datasets and prompts.}
We evaluate on CelebA-HQ \citep{karras2018progressive} and VGGFace2 \citep{cao2018vggface2}, using 50 identities and one source image per identity. Each image is evaluated under 12 instructions covering facial expressions, local attributes, and background changes. Prompts 1--8 form the optimization pool for \method: smiling, anger, sadness, surprise, glasses, beard, beach, and forest. 
Prompts 9--12, including worried expression, hat, smoky-eye makeup, and city street, are excluded from optimization and used only to evaluate cross-prompt generalization.

\paragraph{Target models.}
We evaluate frozen Step1X-Edit \citep{liu2025step1x} and Qwen-Image-Edit-2511 \citep{qwenimage2025}, two VLM-conditioned DiT editors with different conditioning interfaces and generation backbones.

\paragraph{Baselines.}
Unprotected \emph{Clean} editing serves as the normal-edit reference. \emph{DiffPGD} represents denoising-loss attacks \citep{liang2023advdm}; \emph{VAE} represents latent-disruption methods such as PhotoGuard-Encoder and EditShield \citep{salman2023photoguard,chen2024editshield}; and \emph{DeContext} suppresses reference propagation through multimodal attention \citep{shen2025decontext}.

\paragraph{Implementation details.}
Unless otherwise specified, \method uses $\epsilon=16/255$, 1,000 optimization steps, an initial step size of $1/255$ with cosine decay, and momentum $\mu=0.9$. All methods use matched random seeds and official inference configurations. Experiments run on NVIDIA A100 80GB GPUs. Full prompts, model configurations, and baseline objectives are provided in Appendix~D.

\paragraph{Evaluation metrics.}
We report EditReward-MiMo instruction following (IF) \citep{wu2026editreward}, RetinaFace detection failure rate (FDFR) \citep{deng2020retinaface}, ArcFace identity similarity (ISM) \citep{deng2019arcface}, and CLIP-I and SSIM for source fidelity \citep{radford2021clip,wang2004ssim}.

\begin{figure*}[!t]
\centering
\includegraphics[width=\textwidth]{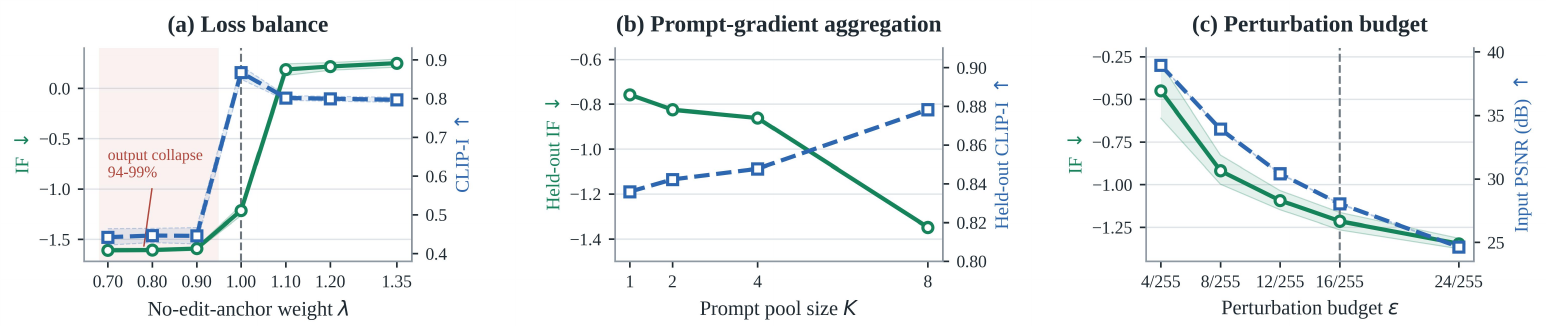}
\caption{Ablations on CelebA-HQ with Step1X-Edit. \textbf{Green} and \textbf{blue} curves report edit suppression and fidelity.}
\label{fig:ablation-sensitivity}
\end{figure*}

\subsection{Main Results}

Table~\ref{tab:main-results} therefore evaluates IF jointly with face validity and source fidelity, while Figure~~\ref{fig:qualitative-main} provides a qualitative comparison of the resulting edits across methods and prompts.

\paragraph{Step1X-Edit.} \method reaches IF values of $-1.218/-1.178$ on CelebA-HQ/VGGFace2, compared with $-0.482/-0.404$ for the SOTA baseline, DiffPGD. Given this edit suppression, \method still achieves the highest protected CLIP-I ($0.867/0.895$) and SSIM ($0.780/0.784$), together with near-zero FDFR ($0.000/0.005$). The outputs therefore retain detectable faces and source semantics while remaining structurally closest to the source.

\paragraph{Qwen-Image-Edit.} 

\method obtains IF values of $-0.897/-0.900$ on CelebA-HQ/VGGFace2, substantially below the SOTA baseline, DiffPGD ($-0.035/-0.022$). DeContext and VAE retain IF scores close to Clean; their near-Clean source-fidelity scores therefore reflect ineffective protection rather than successful stealthy no-op protection. 
Face validity and source fidelity are therefore meaningful for stealthy no-op only when interpreted jointly with sufficiently low IF.
Under this joint criterion, \method combines the lowest IF with the highest protected CLIP-I ($0.801/0.843$) and SSIM ($0.576/0.670$).

\subsection{Prompt Generalization}
\label{sec:prompt-generalization}

\method is optimized with prompts 1--8 and evaluated on four held-out prompts that never participate in perturbation optimization. Table~\ref{tab:prompt-results} shows a representative Step1X-Edit result. Across all settings, \method consistently lowers IF on held-out prompts. To account for differences in prompt difficulty, we measure the IF reduction from Clean to \method within each split. Averaged over CelebA-HQ and VGGFace2, held-out prompts retain $91.2\%$ of the in-pool suppression gain on Step1X-Edit and $94.5\%$ on Qwen-Image-Edit. These results demonstrate cross-prompt transfer beyond the prompt strings used during optimization.
The complete comparison is provided in Appendix~D.

\subsection{Time and Resource Cost}

Table~\ref{tab:resource-cost} reports prompt-normalized GPU cost and peak VRAM. \method consumes fewer GPU-hours and peak VRAM per image than DeContext and DiffPGD. VAE is faster and lighter but provides substantially weaker protection (Table~\ref{tab:main-results}).  More details are provided in Appendix~D.

\subsection{User Study}
\label{sec:perceptual-pilot}
\begin{figure}[t]
\centering
\includegraphics[width=\linewidth]{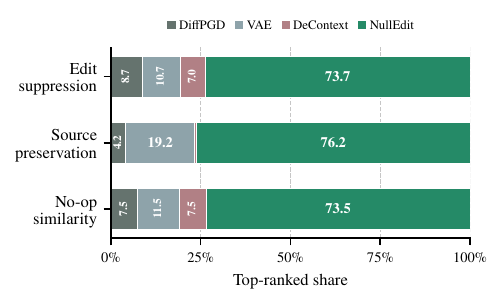}
\caption{User study. Each stacked bar reports the percentage of trials in which each method received the highest score for the corresponding dimension.}
\label{fig:perceptual-pilot}
\end{figure}

We conduct a controlled user study with 10 participants. We assign one prompt to each of the 50 CelebA-HQ identities, approximately balancing prompt frequencies across identities.
For every trial, outputs are anonymously shuffled and scored from 1 to 5 for edit suppression, source preservation, and no-op similarity. 
Figure~\ref{fig:perceptual-pilot} shows that \method receives the top rank in $73.7\%$, $76.2\%$, and $73.5\%$ of trials.
More details are provided in Appendix D.


\begin{table}[!t]
\centering
\small
\setlength{\tabcolsep}{1.6pt}
\begin{tabular}{llccccc}
\toprule
\textbf{Split} & \textbf{Method} & IF$\downarrow$ & FDFR$\downarrow$ &
ISM$\uparrow$ & CLIP-I$\uparrow$ & SSIM$\uparrow$ \\
\midrule
In-pool & Clean
& 0.733 & 0.003 & 0.416 & 0.801 & 0.747 \\
& \textbf{\method}
& \textbf{-1.150} & \textbf{0.000} & \textbf{0.544}
& \textbf{0.862} & \textbf{0.780} \\
\addlinespace[1pt]
\rowcolor{heldoutgray}
Held-out & Clean
& 0.310 & 0.000 & 0.535 & 0.872 & 0.751 \\
\rowcolor{heldoutgray}
& \textbf{\method}
& \textbf{-1.353} & 0.000 & \textbf{0.562}
& \textbf{0.878} & \textbf{0.781} \\
\bottomrule
\end{tabular}
\caption{Cross-prompt transfer on CelebA-HQ using Step1X-Edit. In-pool prompts are used for optimization, whereas held-out prompts are used only for evaluation.}
\label{tab:prompt-results}
\end{table}

\begin{table}[!t]
\centering
\small
\setlength{\tabcolsep}{5.0pt}
\begin{tabular}{lcccc}
\toprule
\textbf{Metric} &
\textbf{DiffPGD} &
\textbf{VAE} &
\textbf{DeContext} &
\textbf{\method} \\
\midrule
GPU-h $\downarrow$
& 0.650 & 0.032 & 0.745 & 0.080 \\
VRAM (GiB) $\downarrow$
& 45.72 & 4.93 & 50.21 & 20.50 \\
\bottomrule
\end{tabular}
\caption{Average per-image protection cost.}
\label{tab:resource-cost}
\end{table}

\subsection{Ablation Studies}

We conduct ablations on CelebA-HQ using Step1X-Edit as the target editor. Figure~\ref{fig:ablation-sensitivity} reports loss balance, prompt-gradient aggregation, and perturbation-budget sensitivity in panels (a)--(c).

\paragraph{Loss balance.}
Varying the no-edit anchor weight reveals a narrow operating point. Values below one yield lower IF ($-1.610$ to $-1.594$), but at the cost of output collapse: FDFR rises to $93.5$ to $99.0\%$, and CLIP-I falls to approximately $0.44$. Increasing $\lambda$ above one preserves valid outputs but sharply weakens edit suppression (IF $0.188$ to $0.251$). Thus, the balanced objective achieves source-preserving edit suppression only near equal weighting.

\paragraph{Prompt-gradient aggregation.}
We use nested prompt pools with $K\in\{1,2,4,8\}$ and evaluate held-out prompts 9--12. Increasing $K$ from one to eight lowers IF from $-0.758$ to $-1.350$ while raising CLIP-I from $0.836$ to $0.878$, showing that a broader representative pool improves transfer to held-out prompts. Consistent with the two-dataset averages reported in the \emph{Prompt Generalization} section, held-out prompts retain over 90\% of the in-pool edit-suppression gain when using the compact prompt pool.

\paragraph{Perturbation budget.}
We vary $\epsilon\in\{4,8,12,16,24\}/255$. IF decreases monotonically from $-0.450$ to $-1.346$, while protected-input PSNR falls from $38.95$ to $24.64$ dB. The default $16/255$ budget provides a practical operating point and is comparable in scale to perturbation budgets commonly used by related baselines.

\section{Conclusion}

We introduced \method, a proactive protection method against unauthorized editing by VLM-conditioned DiT editors. Through balanced VLM condition redirection and cross-prompt gradient averaging, \method turns requested manipulations into stealthy, source-preserving no-ops without relying on visible corruption or identity drift. More broadly, our work identifies the VLM-derived semantic condition as a new intervention surface and reframes proactive image protection from degrading generated outputs to preventing unsafe editing behavior itself, providing a foundation for less observable protection and harmless outputs in future multimodal editors.


\bibliography{references}

\end{document}